\pdfoutput=1

\documentclass[11pt]{article}

\usepackage[]{acl}

\usepackage{times}
\usepackage{latexsym}

\usepackage[T1]{fontenc}

\usepackage[utf8]{inputenc}

\usepackage{microtype}

\usepackage{inconsolata}
\usepackage{graphicx}
\usepackage{xcolor}
\usepackage{amsmath}
\usepackage{makecell}
\usepackage{subcaption}

\title{EMONA: \underline{E}vent-level \underline{M}oral \underline{O}pinions in \underline{N}ews \underline{A}rticles}

\author{Yuanyuan Lei, Md Messal Monem Miah, Ayesha Qamar, Sai Ramana Reddy, \\
        \vspace{3pt}
        {\bf Jonathan Tong, Haotian Xu, Ruihong Huang} \\
        Department of Computer Science and Engineering\\
        Texas A\&M University, College Station, TX\\
        \texttt{\{yuanyuan, huangrh\}@tamu.edu}}

\begin{document}
\maketitle
\begin{abstract}

Most previous research on moral frames has focused on social media short texts, little work has explored moral sentiment within news articles. In news articles, authors often express their opinions or political stance through moral judgment towards events, specifically whether the event is right or wrong according to social moral rules. This paper initiates a new task to understand moral opinions towards events in news articles. We have created a new dataset, \textbf{EMONA}\footnote{https://github.com/yuanyuanlei-nlp/EMONA\_dataset}, and annotated \underline{\bf{e}}vent-level \underline{\bf{m}}oral \underline{\bf{o}}pinions in \underline{\bf{n}}ews \underline{\bf{a}}rticles. This dataset consists of 400 news articles containing over 10k sentences and 45k events, among which 9,613 events received moral foundation labels. Extracting event morality is a challenging task, as moral judgment towards events can be very implicit. Baseline models were built for event moral identification and classification. In addition, we also conduct extrinsic evaluations to integrate event-level moral opinions into three downstream tasks. The statistical analysis and experiments show that moral opinions of events can serve as informative features for identifying ideological bias or subjective events.

\end{abstract}

\section{Introduction}

\begin{figure*}[ht]
  \centering
  \includegraphics[width = 6.3in]{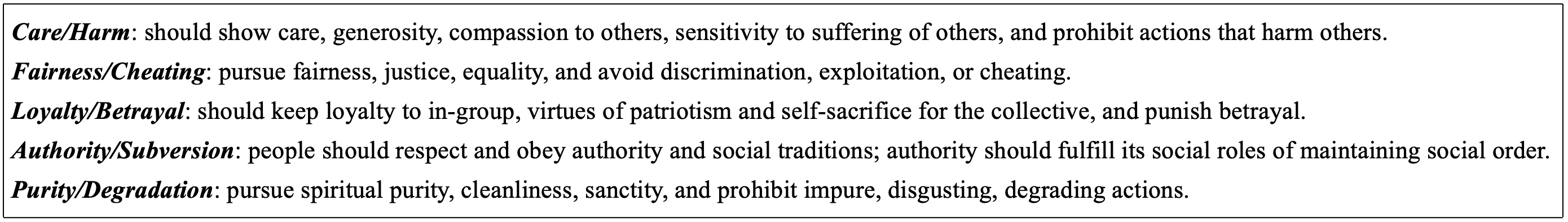}
  \caption{\textit{Moral Foundation Theory} categorizes moral principles into five dimensions.}
  \label{MFT}
\end{figure*}

\begin{figure*}[ht]
  \centering
  \includegraphics[width = 6.3in]{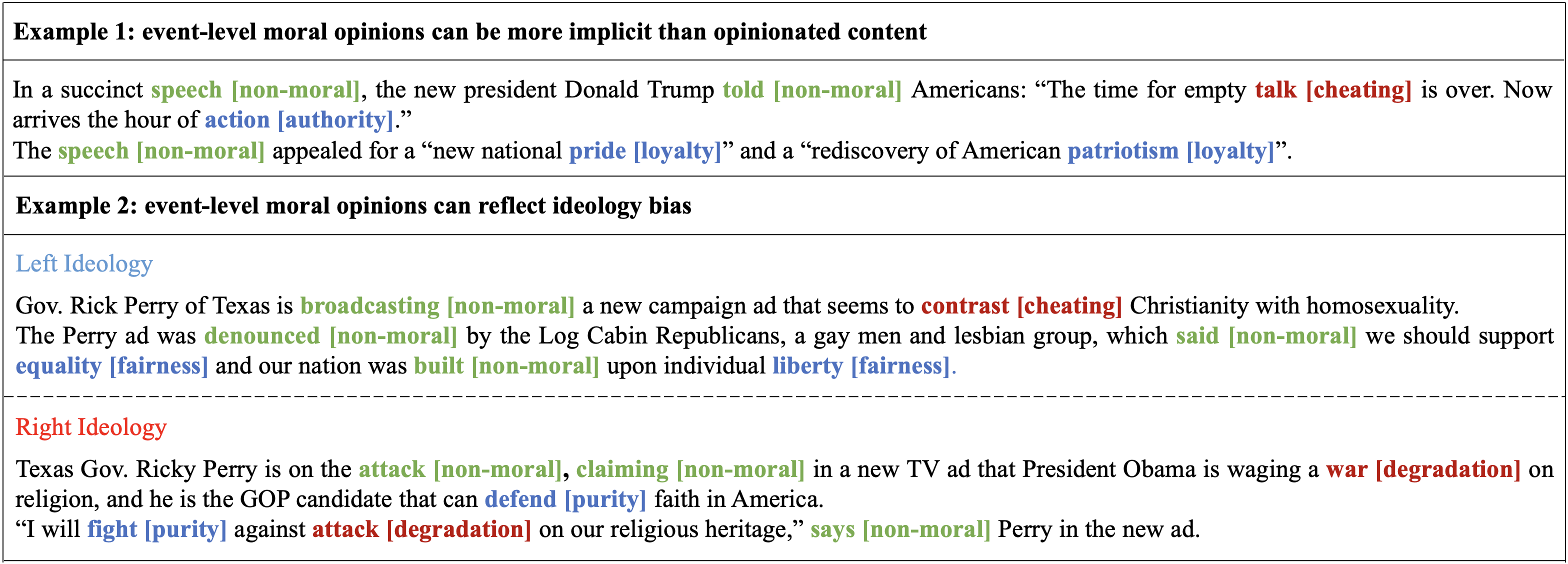}
  \caption{Examples of moral opinions towards events in EMONA dataset.}
  \label{introduction_example}
\end{figure*}

Morality refers to a set of social moral principles to distinguish between right and wrong \cite{berker2019explanatory, nilsson2020moral}. Moral judgment plays a crucial role in expressing public opinions, driving social movements, and shaping policy decisions. \cite{dehghani2016purity, wolsko2017expanding, brady2020mad, voelkel2022changing}. \textit{Moral Foundations Theory} \cite{haidt2007morality, graham2009liberals}, provides a theoretical framework to categorize social moral principles into five dimensions, each associated with a positive and negative judgment: \textit{Care/Harm}, \textit{Fairness/Cheating}, \textit{Loyalty/Betrayal}, \textit{Authority/Subversion}, and \textit{Purity/Degradation} (Figure \ref{MFT} provides detailed values). Extracting moral framing from text demands a combination of sociological moral knowledge and contextual semantic understanding \cite{fulgoni-etal-2016-empirical, xie-etal-2019-text, johnson-goldwasser-2018-classification}.

In many studies, the moral foundations lexicon \cite{Frimer_2019} has been widely used to match words and identify moral foundations in text. Realizing the limitations of this lexicon match approach, researchers have started creating moral foundation annotations in text and training moral foundation detectors using annotations. However, such resource creation efforts have mainly been devoted to  social media short text analysis and moral frames were usually annotated for an entire social media post \cite{trager2022moral, roy-etal-2021-identifying, hoover2020moral}, in contrast, little work has explored moral sentiment within news articles at a more fine-grained and nuanced level yet.

In this paper, we propose a new task to understand moral opinions towards events in news articles. The concept of \textit{event} refers to an occurrence or action, and is the basic element in story telling \cite{zhang-etal-2021-salience, lei-huang-2023-identifying}. In news media, the authors often express their stance through moral judgment towards events, so as to shape public opinions \cite{wolsko2016red, amin2017association, lei-huang-2022-shot}. To facilitate a profound study towards morality aspect of events, we create a new dataset, EMONA, annotated with Event-level Moral Opinions in News Articles.

While we believe the created dataset EMONA can be widely useful for studying moral opinion injection in context, this effort is initially motivated by the potential significant role of moral opinions for media bias analysis. In addition, recognizing the difficulty of identifying subjective events in news articles, we also aim to understand whether event moral opinions enables uncovering implicit subjective events that are otherwise hard to detect. To address these research questions, we have carefully chosen 400 documents from three sources for annotation, including 180 documents from AllSides that spans over 12 domains and indicates article-level ideology bias \cite{baly-etal-2020-detect}; half (150 documents) of the BASIL dataset that has sentence-level media bias annotated \cite{fan-etal-2019-plain}; and the entire MPQA 3.0 dataset (70 documents) that has fine-grained opinions toward entities and events annotated \cite{deng-wiebe-2015-mpqa}.

The annotation process went through two passes: the first pass annotates event mentions, and the second pass annotates moral opinions for individual event mentions with respect to the context of the news article. Annotating event-level moral opinions within a news article turned out to be a challenging and demanding task for human annotators,  we recruited five annotators and steadily improved pairwise inter-annotator agreements for both tasks to a satisfactory level. In total, the dataset contains over 10k sentences and 45k event mentions, among which, 9,613 event mentions were annotated as bearing moral opinions.

The challenge of event morality analysis indeed lies in the implicit nature of moral opinions toward events. In some cases, event-level moral opinions can be too implicit to be identified as opinions. Take the example 1 in Figure \ref{introduction_example} as an instance, which are annotated as non-opinionated content by previous work \cite{fan-etal-2019-plain, deng-wiebe-2015-mpqa}, the semantic implies \textit{cheating} criticism towards \textit{empty talk}, and infers praise towards \textit{patriotism} event. The implicit nature of event-level moral opinions requires the annotators to take both local sentential context and global broader context into consideration during annotations.

The numerical and visualization analysis of the dataset show that annotated event-level moral opinions can effectively reflect article-level ideology, and designate sentence-level political bias. Take the example 2 in Figure \ref{introduction_example} as an illustration, liberal media focus on \textit{fairness} judgment by framing the story as \textit{Texas Gov. contrasts Christianity with homosexuality}, while conservative media emphasizes on \textit{purity} moral by praising \textit{Texas Gov. defends religious faith}. While not evident to readers, the journalists often subtly influence public opinions through moral value implication \cite{roy-goldwasser-2021-analysis}. The annotated moral opinions towards events can uncover and provide fine-grained explanations for such ideological bias.

We build baseline models for event moral identification and classification, and we further conduct extrinsic evaluations on three downstream tasks: article-level ideology classification, sentence-level media bias identification, and event-level opinions identification. The experiments demonstrate the usefulness of detecting event-level moral opinions on all the three extrinsic evaluation tasks, with F1 score improved by 3.35\% to 4.71\%, and also validate the value of our new dataset.


\section{Related Work}

\textbf{Moral Foundation Theory} is a theoretical framework developed by sociologists \cite{haidt2004ethics, haidt2007morality, graham2009liberals} which categorizes moral values into five primary dimensions. It has been extensively employed to examine the influence of moral principles on human behaviors, judgments, decision-making, and political stances \cite{voelkel2022changing, hoover2018moral, lei-huang-2023-discourse, brady2017emotion, 10008998, fulgoni-etal-2016-empirical}.

\noindent \textbf{Moral Foundations Lexicon} \cite{Frimer_2019}  was widely employed in many applicational studies \cite{ramezani-etal-2021-unsupervised-framework, field-etal-2018-framing, garten2016morality} for identifying moral foundations. However, this lexicon match approach suffers from both over-labeling commonly used indicator words and overlooking context-dependent expressions of moral foundations. Researchers \cite{ramezani-etal-2021-unsupervised-framework} have considered leveraging distributional word embeddings to mitigate the drawbacks of using the Lexicon only.

\noindent \textbf{Annotating Moral Frames for Social Media} moral frame annotation efforts have mainly been devoted to social media short text analysis and moral frames have been annotated on Twitter microblogs \cite{johnson-goldwasser-2019-modeling,hoover2020moral} and Reddit posts \cite{trager2022moral}. Moral frames were usually annotated for an entire social media posts, \citet{roy-etal-2021-identifying} further annotates moral roles (individuals or collective enties) for a moral frame. \citet{shahid-etal-2020-detecting} is the only prior work we are aware of that annotates moral foundations for news articles, but their annotations were conducted at the sentence level.

\noindent \textbf{Event Subjectivities:} a body of prior research have dedicated to studying the subjective aspects of meanings for events. \citet{ding-riloff-2018-human, zhuang-etal-2020-affective, zhuang-riloff-2023-eliciting}  acquired subjective events from texts and recognizing their polarities (\textit{positive} or \textit{negative}). \citet{rashkin-etal-2016-connotation,feng-etal-2013-connotation} acquired connotation lexicons and connotation frames for event predicates. \citet{deng-wiebe-2015-mpqa} annotates and studies events as opinion targets in the context of a news article. 

\section{EMONA Dataset Construction}

\begin{table*}[ht]
    \centering
    \scalebox{0.89}{
    \begin{tabular}{|l|ccc|ccc||c|c|c|}
        \hline
        & \multicolumn{3}{c|}{Pairwise} & \multicolumn{3}{c||}{Majority} & llama-2 & gpt-3.5 & gpt-4 \\
        \hline
        & Min & Max & Mean & Min & Max & Mean & Mean & Mean & Mean \\
        \hline
        Event Extraction & 0.7670 & 0.8182 & \textbf{0.7938} & 0.8656 & 0.8877 & 0.8772 & 0.4464 & 0.5870 & 0.7113 \\
        Moral Identification & 0.4797 & 0.5846 & \textbf{0.5276} & 0.6483 & 0.7814 & 0.7017 & 0.3374 & 0.4064 & 0.4707 \\
        Ten Moral Classification & 0.6682 & 0.7876 & \textbf{0.7230} & 0.8279 & 0.9245 & 0.8679 & 0.3685 & 0.4068 & 0.5546 \\
        \hline
    \end{tabular}}
    \caption{Pairwise Cohen's kappa inter-agreement scores among five annotators, and their agreement with majority voting label. The right three columns show the average agreement scores between each large language model (llama-2-7b-chat, gpt-3.5-turbo, and gpt-4) and the five human annotators.}
    \label{IAA}
\end{table*}

\begin{table}[ht]
    \centering
    \scalebox{0.83}{
    \begin{tabular}{|c|cccc|}
        \hline
        & \# Doc & \# Sent & \# Event & \# (\%) Moral \\
        \hline
        ALL & 400 & 10912  & 45199  & 9613 (21.27) \\
        \hline
        AllSides & 180 & 5448  & 22527  & 5628 (24.98)\\
        BASIL & 150 & 3811  & 16200  & 2586 (15.96) \\
        MPQA 3.0 & 70 & 1653  & 6472  & 1399 (21.62) \\
        \hline
    \end{tabular}}
    \caption{Statistics of EMONA and its three portions. \% Moral represents the ratio of moral related events in all the annotated events.}
    \label{dataset_statistics_1}
\end{table}

\begin{table*}[ht]
    \centering
    \scalebox{0.67}{
    \begin{tabular}{|c|cccccccccc|}
        \hline
        & \textcolor{blue}{Care} & \textcolor{red}{Harm} & \textcolor{blue}{Fairnss} & \textcolor{red}{Cheating} & \textcolor{blue}{Loyalty} & \textcolor{red}{Betrayal} & \textcolor{blue}{Authority} & \textcolor{red}{Subversion} & \textcolor{blue}{Purity} & \textcolor{red}{Degradation} \\
        \hline
        ALL & 518 (5.39) & 1815 (18.88) & 688 (7.16) & 2445 (25.43) & 99 (1.03) & 97 (1.01) & 1252 (13.02) & 2510 (26.11) & 90 (0.94) & 99 (1.03) \\
        \hline
        AllSides & 345 (6.13) & 1367 (24.29) & 414 (7.36) & 1435 (25.50) & 50 (0.89) & 60 (1.07) & 496 (8.81) & 1384 (24.59) & 50 (0.89) & 27 (0.48) \\
        BASIL & 104 (4.02) & 271 (10.48) & 187 (7.23) & 738 (28.54) & 42 (1.62) & 32 (1.24) & 471 (18.21) & 678 (26.22) & 25 (0.97) & 38 (1.47) \\
        MPQA 3.0 & 69 (4.93) & 177 (12.65) & 87 (6.22) & 272 (19.44) & 7 (0.50) & 5 (0.36) & 285 (20.37) & 448 (32.02) & 15 (1.07) & 34 (2.43) \\
        \hline
    \end{tabular}}
    \caption{Number (Ratio) of events with ten moral labels in the EMONA dataset and its three portions. Blue moral labels represent \textcolor{blue}{positive} moral opinions, and red moral labels represent \textcolor{red}{negative} moral opinions.}
    \label{dataset_statistics_2}
\end{table*}


\subsection{Data Sources}

We intentionally select three different components to form our dataset. Overall, the dataset comprises 400 news articles, 10k sentences and 283k words.

\begin{itemize}
    \item \textit{AllSides} \cite{baly-etal-2020-detect} provides article-level ideology stance labels: \textit{left}, \textit{center}, and \textit{right}. We select 12 frequently discussed topics: \textit{abortion}, \textit{coronavirus}, \textit{elections}, \textit{gun control}, \textit{immigration}, \textit{lgbt rights}, \textit{race and racism}, \textit{violence}, \textit{politics}, \textit{us house}, \textit{us senate}, \textit{white house}. Each topic comprises five articles for each of the three stances, spanning from 2012 to 2020, resulting in a total of 180 articles.
    \item \textit{BASIL} \cite{fan-etal-2019-plain} provides sentence-level media bias labels and collects 100 triples of articles from three medias (Fox News, New York Times, Huffington Post) discussing the same event. We sample five triples for each year from 2010 to 2019, leading to 50 articles from each media and a total of 150 articles.
    \item \textit{MPQA 3.0} \cite{deng-wiebe-2015-mpqa} annotates general opinions towards entities and events for 70 articles from the years 2001 to 2002. We retained all 70 articles for the study of the correlation between event-level moral opinions and opinionated events.
\end{itemize}

\subsection{The Annotation Procedure}

The annotators were instructed to annotate a plain article in two passes, identifying event mentions and assigning moral labels for events. Before any annotation takes place, we ask our annotators to first read the entire article and understand the author’s overall judgement, stance and opinions. Then in the first pass of annotations, our annotators identify all the event mentions in a news article and annotate individual words as event mentions, following the annotation guidelines of Richer Event Description \cite{ogorman-etal-2016-richer} in identifying minimum spans of event mentions. In the second pass of annotations, our annotators examine each event mention with respect to its local and global contexts, and identify if the event bears any moral judgement from the author or the source for events appearing in quotations. For the event mentions that carry moral judgements, the annotators will assign one moral dimension to each event mention, the primary dimension out of the ten moral dimensions. Even though we do not explicitly annotate event arguments \footnote{Annotating event arguments itself is a strenuous task and event extraction is not the main focus of creating this dataset.}, we ask our annotators to identify the agent and patient of each event mention and consider the intention of the agent and the effect on the patient when determining moral judgements toward an event.

We recruited five annotators\footnote{Four graduate students and one undergraduate student conducting research in natural language processing} and conducted annotation training until a satisfactory level of inter-annotator agreement was reached. 
Identifying event mentions (the first pass) is relatively straightforward, but recognizing and classifying moral opinions in context (the second pass) has been approved to be difficult and takes rounds of discussions to reach a stable level of agreement among our multiple annotators. In the official annotation process, we first asked all of our annotators to annotate a common set of 25 documents, and then evenly distributed the remaining documents among our annotators and had each annotator label another 75 documents.

\subsection{Inter-annotator Agreements (IAAs)}

Based on the common set of 25 documents that contains 3,194 events, we calculate Cohen's kappa inter-annotator agreements (IAAs) between each pair of annotators.  In Table \ref{IAA} (the first column), we report the minimum, maximum, and mean of the pairwise agreement scores among five annotators for each of the three tasks, event identification (identifying each word as event or not event), moral identification (identifying an event as moral or non-moral), and ten moral classification (classifying among the ten moral dimensions for events that have been identified as moral). We also report agreements between each annotator and the majority voting labels (the second column). We can see that identifying events is a relatively easy task, and identifying whether an event bears any moral judgment turns out to be more difficult than discerning among the ten moral dimensions.


\begin{figure*}[ht]
  \centering
  \includegraphics[width = 6.3in]{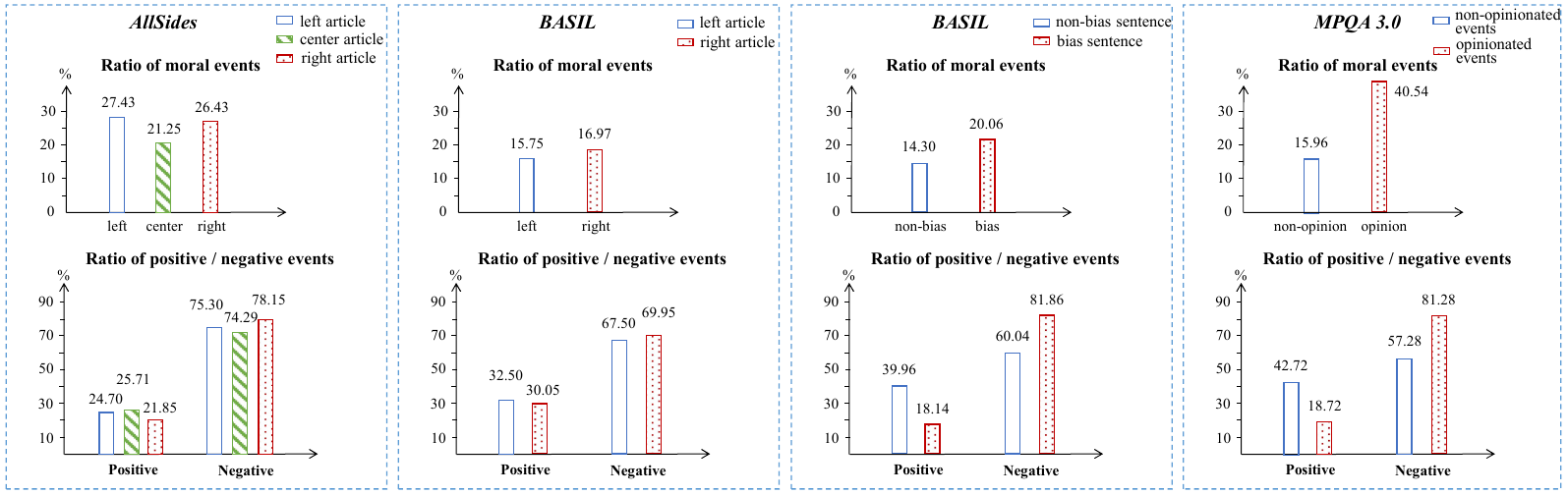}
  \caption{The ratio of moral events in all the annotated events, as well as the ratio of positive / negative events in moral events in different portions of EMONA dataset.}
  \label{analysis_figure_1}
\end{figure*}

\subsection{Large Language Models vs. Humans}

We further investigate the capabilities of large language models  \cite{openai2023gpt4, touvron2023llama} in event morality annotation. The large language models gpt-4, gpt-3.5-turbo, and llama-2-7b-chat were prompted to automatically identify events and assign event-level moral labels for the same set of articles (the used prompt is in Appendix \ref{prompt}). The agreement scores between model generated labels and human annotators are presented in the right three columns of Table \ref{IAA}. Compared to the other two large language models, gpt-4 aligns better with human annotators on event identification and event morality identification and classification. But the model vs. human agreement levels are overall still lower than human vs. human agreements, especially in classifying among the ten moral dimensions.  

\subsection{Dataset Statistics}

Table \ref{dataset_statistics_1} shows the basic statistics of the EMONA dataset and its three portions. In total, 45,199 event mentions were annotated in EMONA and roughly 21 percents of them (9,613 event mentions) were assigned with a moral label. Table \ref{dataset_statistics_2} further shows the distributions of the ten moral labels. We can see that \textit{Care/Harm}, \textit{Fairness/Cheating} and \textit{Authority/Subversion} are much more frequent than the other two dimensions \textit{Loyalty/Betrayal} and \textit{Purity/Degradation}. Within each of the three most common moral dimensions, the negative class is two or three times more frequent than the positive class. The imbalanced distribution of the ten moral classes poses a challenge for automatic event morality analysis.


\section{Dataset Analysis}

This section provides a numerical and visual analysis (Figure \ref{analysis_figure_1}, \ref{analysis_figure_2}) of the correlations between event-level moral opinions and article-level ideology, sentence-level media bias, and event-level general opinions.

\subsection{Analysis with Article-level Ideology}

We aim to address the research question of whether event-level moral opinions can reflect article-level ideology. The ratios of moral events across articles from left, center, and right media outlets within the AllSides portion are plotted in the first subfigure of Figure \ref{analysis_figure_1}. We observe that the articles with center stance have a lower ratio of moral events than the articles with polarized stances. The left or right articles exhibit a higher ratio of negative moral events, while center articles carry more positive moral judgments. In the second subfigure of Figure \ref{analysis_figure_1}, we also plot the ratio of moral events in the left leaning and right leaning articles within the BASIL portion, where left articles are from Huffington Post and New York Times while right articles are from Fox News. The detailed ratios of moral events distributed among ten moral foundation categories can be found in Appendix \ref{appendixB}. We can see that the articles with different ideologies showcase different preferences of moral foundations. The analysis validates that our event-level moral opinions annotations can inform article-level ideological bias.

\subsection{Analysis with Sentence-level Media Bias}

We also explore the research question of whether event-level moral opinions can designate sentence-level political bias. The ratios of moral events across non-bias and bias sentences in the BASIL portion are plotted in the third subfigure of Figure \ref{analysis_figure_1}. 
It is evident that bias sentences exhibit a higher ratio of moral events than non-bias sentences. Moreover, bias sentences tend to include more negative moral judgments, indicating that the journalists often influence public opinions through moral criticism. In contrast, non-bias sentences display negative criticism and positive praise more evenly. The analysis confirms that moral opinions towards events can designate and explain sentence-level media bias.

\begin{table*}[ht]
    \centering
    \scalebox{0.9}{
    \begin{tabular}{|l|ccc|ccc|ccc|}
        \hline
        & \multicolumn{3}{c|}{event moral identification} & \multicolumn{3}{c|}{event moral classification} & \multicolumn{3}{c|}{end-to-end system} \\
        & Precision & Recall & F1 & Precision & Recall & F1 & Precision & Recall & F1 \\
        \hline
        lexicon & 36.49 & 27.80 & 31.56 & 18.21 & 19.39 & 16.75 & - & - & - \\
        gpt-3.5-turbo & 41.45 & 58.90 & 48.66 & 24.20 & 27.46 & 22.08 & 22.98 & 23.14 & 20.61 \\
        gpt-4 & 59.25 & 60.91 & 60.06 & 35.08 & 32.68 & 30.83 & 30.64 & 32.73 & 30.06 \\
        longformer & 61.81 & 65.48 & 63.59 & 46.32 & 36.56 & 39.50 & 44.30 & 35.47 & 38.42 \\
        \hline
    \end{tabular}}
    \caption{Performance of intrinsic evaluations on EMONA dataset. The last row represents the results of our built evaluation model. Ten-folder cross validation is conducted.}
    \label{intrinsic_results}
\end{table*}

\subsection{Analysis with Event-level Opinions}

The relation between event-level moral opinions and event-level general opinions is another research question. The fourth subfigure of Figure \ref{analysis_figure_1} presents the ratios of moral events among the opinionated / non-opinionated events annotated in the MPQA 3.0 dataset. The opinionated events indeed carry a higher ratio of moral judgments, especially negative criticism. The analysis proves the positive correlation between moral judgments and general opinions towards events.

In addition, we also look into the difference between moral events and opinionated events, and show the confusion matrix in Table \ref{analysis_with_opinions}. There are 886 (59.46\%) opinionated events not relevant to moral judgments, which implies that not all opinions are moral opinions. For example, \textit{the medical policy hurts the economy}, where the author conveys negative sentiment towards \textit{hurt} event but the judgment is evaluative and not from the moral perspective. Also, we observe that 795 (56.82\%) moral events are not identified as opinionated events. This indicates that event-level moral opinions can be more implicit than opinionated content, thereby can supplement implicit opinions. Hence, our event-level moral opinions annotations can uncover and supplement general opinions.

\begin{table}[ht]
    \centering
    \scalebox{0.8}{
    \begin{tabular}{|c|cc|}
        \hline
        & moral events & non-moral events \\
        \hline
        opinionated events & 604 & 886 \\
        non-opinionated events & 795 & 4147 \\
        \hline
    \end{tabular}}
    \caption{Confusion matrix of the number of moral events and opinionated events in MPQA 3.0 dataset.}
    \label{analysis_with_opinions}
\end{table}

\vspace{-2pt}

\section{Intrinsic Evaluation}

In this section, we conduct intrinsic evaluations for the newly created dataset EMONA. Specifically, we propose the following tasks and build evaluation models for them: (1) event moral identification: identify whether an event contains moral judgment or not (2) event moral classification: classify moral labels for events, including ten moral labels and \textit{non-moral} label (3) end-to-end system: predict label for every words in a plain article, including ten moral labels, \textit{non-moral} label, and \textit{non-event} label. The former two tasks target towards events, while the end-to-end system generates labels for every words. The intrinsic evaluation models we built can serve as baselines for future work.

\subsection{The Baseline Models}

Considering the news articles are usually long, we utilize the Longformer \cite{beltagy2020longformer} as the language model to encode the entire article. We also add an extra layer of Bi-LSTM \cite{huang2015bidirectional} on top to capture the contextual information. Given an article consisting of $n$ words, the derived word embeddings are $(w_1,\dots,w_n)$, among which the events embeddings are $(e_1,\dots,e_m)$.

The event moral identification model builds a two-layer binary classification head on top of the event embedding $e_i$ to make predictions. The event moral classification model also builds an 11-class classification head on top of the event embeddings $e_i$, to predict the probability of each moral label. The end-to-end system builds a 12-class classification head on top of each word embedding $w_i$ and generates labels for every word. The classical cross-entropy loss is employed for training.

\begin{figure*}[ht]
  \centering
  \includegraphics[width = 6.3in]{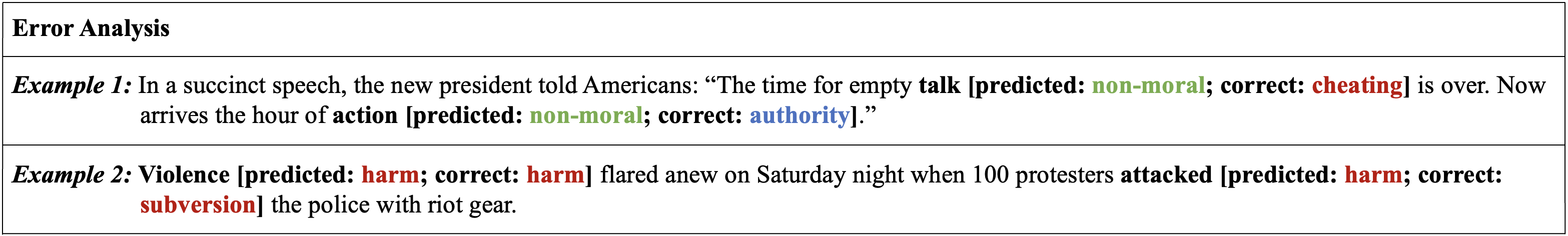}
  \caption{Error analysis of event-level moral opinions classification}
  \label{error_analysis}
\end{figure*}

\subsection{Experimental Settings}

Ten-fold cross validation is performed for evaluation. The entire dataset EMONA is split into ten folders of the equal size. In each iteration, a fold is used as the test set, eight folds are used as the training set while a remaining fold is used as the validation set to determine when to stop training. The evaluation metrics are calculated based on all the ten test folders. Precision, Recall, and F1 score of the positive class are reported for event moral identification task. Macro Precision, Recall, and F1 score are reported for event moral classification and the end-to-end system. 

The AdamW \cite{loshchilov2019decoupled} is used as the optimizer. The maximum training epochs is 10. The learning rate is initialized as 1e-5 and adjusted by a linear scheduler. The weight decay is set to 1e-2.

\subsection{Experimental Results}

Table \ref{intrinsic_results} presents the performance of intrinsic evaluations. The last row shows our evaluation models based on longformer. We also implement three systems for comparison: a lexicon matching system, where we match event words with the moral foundations lexicon provided by \citet{Frimer_2019}, and two large language models gpt-3.5-turbo / gpt-4.

The evaluation models based on longformer perform better than the lexicon matching baseline. It is because event-level moral opinions are context-dependent. The same event mentions in different context can take different moral judgments. The simple word-based matching without considering the context cannot well extract these moral opinions. This suggests future directions of encoding broader contextualized information into the model. Also, the two large language models do not surpass the evaluation model based on fine tuning.

The current performance of the end-to-end system is relatively low, and the primary obstacle is the comprehension of event-level moral opinions. The difficulty of moral classification lies in the imbalanced distribution of ten moral classes, which suggests future directions of improving performance on infrequent classes that lack training data.

\subsection{Error Analysis}

In addition, we also perform an error analysis for event-level moral opinions classification. Firstly, we observe that failing to recognize implicit moral opinion is one type of error. Take example 1 in Figure \ref{error_analysis} as an instance, the author implicitly criticizes \textit{empty talk} and subtly praises \textit{action} event. While the model fails to discover the implicitly conveyed opinions and wrongly selects \textit{non-moral} label. This suggests the future directions of incorporating contextual knowledge and enhancing the capability to extract implicit opinions.

\begin{figure}[ht]
  \centering
  \includegraphics[width = 3in]{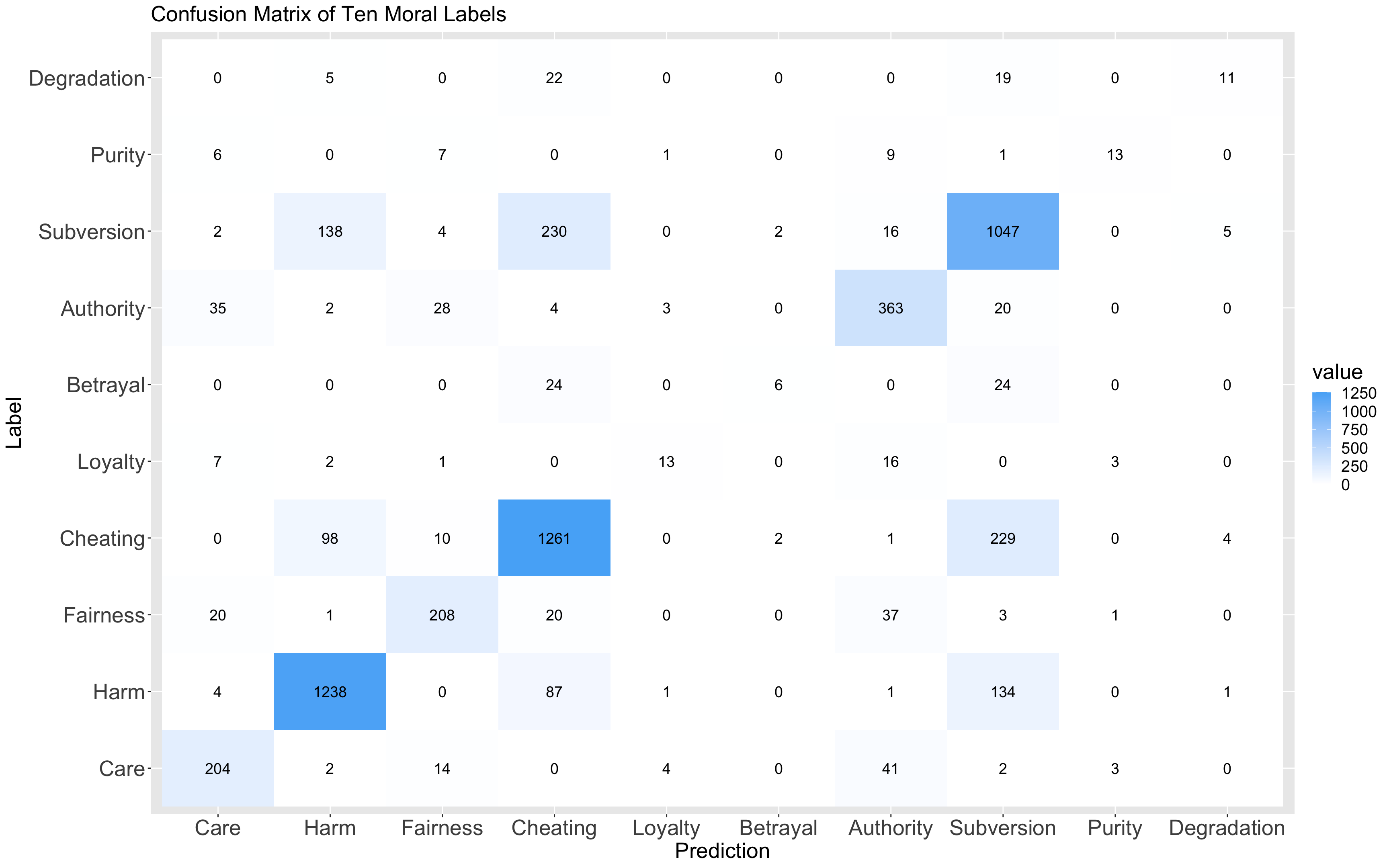}
  \captionsetup{skip = 5pt}
  \caption{Confusion matrix of ten moral labels}
  \label{confusion_matrix}
\end{figure}

Secondly, failing to select the correct moral dimension is another type of error. Figure \ref{confusion_matrix} shows the confusion matrix of ten moral labels. We can see that the confusions between \textit{Harm}, \textit{Cheating}, and \textit{Subversion} are the primary errors. Take example 2 in Figure \ref{error_analysis} as an illustration, the model wrongly predicts \textit{attacked} event as \textit{harm}, by focusing on the semantic of event trigger word, while neglecting the event argument \textit{police} that represents \textit{authority}. This indicates the necessity to encode the model with contextualized semantics.

\section{Extrinsic Evaluation}

\begin{figure*}[ht]
  \centering
  \includegraphics[width = 6.3in]{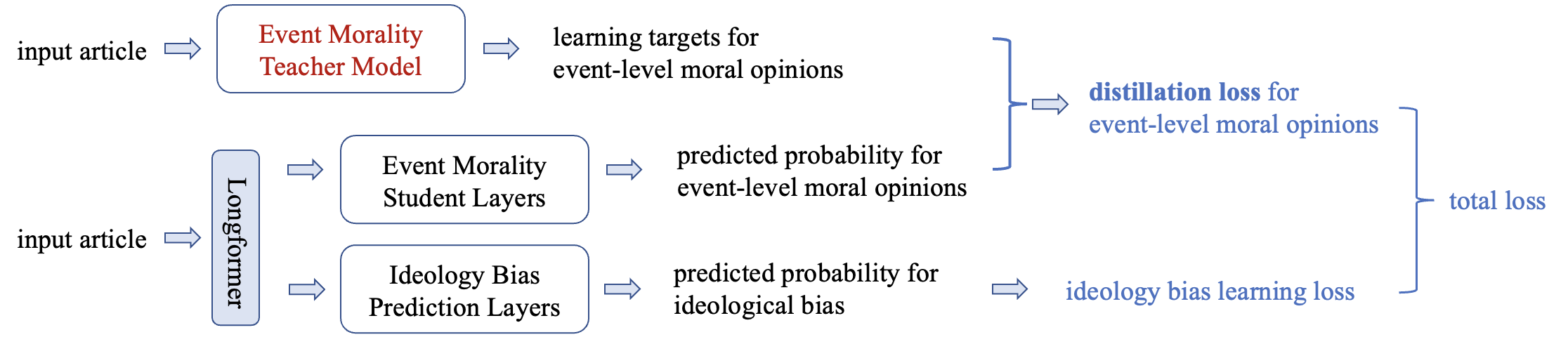}
  \caption{An illustration of ideology bias task informed by event-level moral opinions via knowledge distillation}
  \label{methodology_figure}
\end{figure*}

\begin{table*}[ht]
    \centering
    \scalebox{0.85}{
    \begin{tabular}{|l|ccc|ccc|ccc|}
        \hline
        & \multicolumn{3}{c|}{article-level ideology} & \multicolumn{3}{c|}{sentence-level media bias} & \multicolumn{3}{c|}{event-level opinions} \\
        & Precision & Recall & F1 & Precision & Recall & F1 & Precision & Recall & F1 \\
        \hline
        longformer baseline & 83.69 & 84.65 & 84.11 & 46.81 & 45.65 & 46.22 & 56.73 & 57.71 & 57.21 \\
        + event-level moral opinions & 87.15 & 88.03 & 87.46 & 50.79 & 49.60 & 50.19 & 60.93 & 62.94 & 61.92 \\
        \hline
    \end{tabular}}
    \caption{Performance of article-level ideology classification on AllSides, sentence-level media bias identification on BASIL, and event-level opinions identification on MPQA 3.0 datasets, informed by event-level moral opinions.}
    \label{extrinsic_results}
\end{table*}

To validate the potential applications of this dataset, we further conduct extrinsic evaluations. Motivated by the analysis that confirms positive correlations between moral opinions of events and ideological bias or general opinions, we propose to incorporate event-level moral opinions into three extrinsic evaluation tasks: (1) article-level ideology classification (2) sentence-level media bias identification (3) event-level opinion identification.

\subsection{Knowledge Distillation}

In particular, we design a knowledge distillation framework to distill the event-level moral knowledge into these downstream tasks, as illustrated in Figure \ref{methodology_figure}. This knowledge distillation framework is able to capture moral opinions for articles with or without event morality golden annotations, and integrate the moral knowledge for various downstream extrinsic evaluation tasks.

Specifically, given a new article that does not have moral annotations, the end-to-end event moral classification system is leveraged as the \textit{teacher model} $T_{moral}$, and generates the predicted probability of moral labels for each word $w_i$:
\begin{equation}
    P_i = (p^1_i, p^2_i, ..., p^{12}_i)
\end{equation}
where the predicted probabilities $P_i$ contains the moral knowledge from teacher model $T_{moral}$, and captures the moral opinions for this new article.

The \textit{student layer} of event morality learning is also built on top of word embeddings $w_i$, so as to learn the moral knowledge from the teacher model:
\begin{equation}
    Q_i = softmax(W_2(W_1 w_i + b_1) + b_2)
\end{equation}
where the predicted probabilities $Q_i$ is the learned outcome of student layer.

The \textit{distillation loss} is designed to penalize the distance between the learning targets $P_i$ generated by the teacher model and the learned outcome $Q_i$ of the student layer:
\begin{equation}
    Loss_{moral} = KL(P, Q) = \sum_{i=1}^nP_i log(\frac{P_i}{Q_i})
\end{equation}
where $KL$ means the Kullback-Leibler (KL) divergence loss. By minimizing this distillation loss, the moral knowledge from the teacher can be distilled into the model, and the student layers are forced to be updated with event moral knowledge.

The \textit{task-specific layers} are built on top of the encoder. The article-level ideology prediction layers are built on top of the article start token embedding, to predict article ideology into left, center, and right three classes. The sentence-level media bias prediction layers are built on top of the sentence start token embedding, to predict whether the sentence contains political bias or not. The event-level opinions identification layers are built on top of the event embedding $e_i$, to predict whether the event carries general opinions or not. The classical cross-entropy loss is used to calculate the task-specific loss. The learning objective is the summation of the task-specific loss and the distillation loss.

\subsection{Evaluation Datasets}

The article-level ideology classification is evaluated on AllSides dataset \cite{baly-etal-2020-detect}. We follow the same splitting setting released by the dataset: 27978 train, 6996 valid, and 1300 test articles. The sentence-level media bias identification is evaluated on the entire 300 articles of BASIL dataset \cite{fan-etal-2019-plain}. We follow the previous work \cite{lei-etal-2022-sentence, fan-etal-2019-plain} and conduct ten-folder cross validation for evaluation. The event-level opinions identification is experimented on 70 articles in MPQA 3.0 dataset \cite{deng-wiebe-2015-mpqa}, and we also perform ten-folder cross validation due to its small size.

\subsection{Experimental Results}

The performance of the three extrinsic evaluations are summarized in Table \ref{extrinsic_results}. Macro Precision, Recall, and F1 score are reported for article-level ideology classification task. Precision, Recall, and F1 score of the positive class are reported for the other two binary classification tasks.

Incorporating event-level moral opinions can noticeably improve both precision and recall on the three extrinsic tasks, leading to F1 score increased by 3.35\% to 4.71\%. The probabilistic features provided in the teacher model contains fuzzy and nuanced moral knowledge, thus can benefit the three downstream tasks. The performance gains demonstrate the usefulness of event-level moral opinions, and validates the potential applications of our newly created dataset EMONA.

\section{Conclusion}

This paper defines a new task that understands moral opinions towards events in news articles. We have created a new dataset EMONA, and provide the first annotations of \underline{e}vent-level \underline{m}oral \underline{o}pinions in \underline{n}ews \underline{a}rticles. The dataset analysis, intrinsic evaluation, and extrinsic evaluation on three downstream applications are conducted for this new dataset, which showcase that event-level moral opinions can effectively reflect article-level ideology, designate sentence-level political bias, and uncover event-level implicit opinions. For future work, we will continue to improve the performance of event level moral opinion identification and classification, in addition, we are interested to explore other applications of this task, such as stance detection and news summarization.


\section*{Limitations}

We built the intrinsic evaluation models for event moral identification and classification, along with an end-to-end system. The future work is supposed to explore more sophisticated methodology to identify and classify event-level moral opinions, and incorporate broader contextual information into event morality understanding. Additionally, we propose a knowledge distillation framework to learn ideological bias tasks and event morality task together. Future work necessities the development of more sophisticated methods to incorporate event-level moral opinions with other correlated tasks.

\section*{Acknowledgements}

We would like to thank the anonymous reviewers for their valuable feedback and input. We gratefully acknowledge support from National Science Foundation via the award IIS-2127746. Portions of this research were conducted with the advanced computing resources provided by Texas A\&M High-Performance Research Computing.

\bibliography{anthology,custom}

\appendix

\newpage

\section{Prompt for Large Language Models}

\label{prompt}

The designed prompt to provide large language models with annotation examples is: "An event word is the word describing a thing that happens, such as occurrence, action, process, or event state. Please extract the event words from the sentence, only one word for one event. Please also provide one of the moral labels based on the Moral Foundation Theory (non-moral, care, harm, fairness, cheating, authority, subversion, loyalty, betrayal, sanctity, degradation). If the moral sense is vague, choose 'non-moral'. Please mimic the following examples style. Example: "More than 200 people crowded in the forum on Friday." Event words: "crowded (non-moral)". Example: "We show empathy for other people who might choose abortion." Event words: "empathy (care), choose (non-moral), abortion (non-moral)". Example: "Mayor asked New Yorkers to report after the execution-style killing of officers." Event words: "asked (non-moral), report (non-moral), killing (harm)". Example: "Colorado law prohibits discrimination on the basis of sexual orientation." Event words: "prohibits (fairness), discrimination (cheating)". Example: "People participating in the worship comply with the social distancing orders issued by the government." Event words: "worship (non-moral), comply (authority), issued (non-moral)". Example: "FBI director praised the massive manhunt as an extraordinary effort by law enforcement." Event words: "praised (non-moral), manhunt (non-moral), effort (authority)". Example: "The mobs are overthrowing the government." Event words: "overthrowing (subversion)". Example: "It is the Democrats fault for being weak and ineffective with border security and crime." Event words: "weak (subversion), ineffective (subversion)". Example: "They have fealty and allegiance to out country." Event words: "fealty (loyalty), allegiance (loyalty)". Example: "Trump is accused of colluding with Russia." Event words: "accused (non-moral), colluding (betrayal)". Example: "He calls for putting faith at the center." Event words: "calls (non-moral), faith (sanctity)". Example: "I think anyone who would suggest the military mission is not a success does disservice to the sacrifice of Chief Ryan Owens." Event words: "suggest (non-moral), success (non-moral), disservice (degradation), sacrifice (loyalty)". Sentence: "xxx" Event words:",

\newpage

\section{Distributions Over Ten Moral Foundations}
\label{appendixB}

\begin{figure*}[ht]
  \centering
  \includegraphics[width = 6.3in]{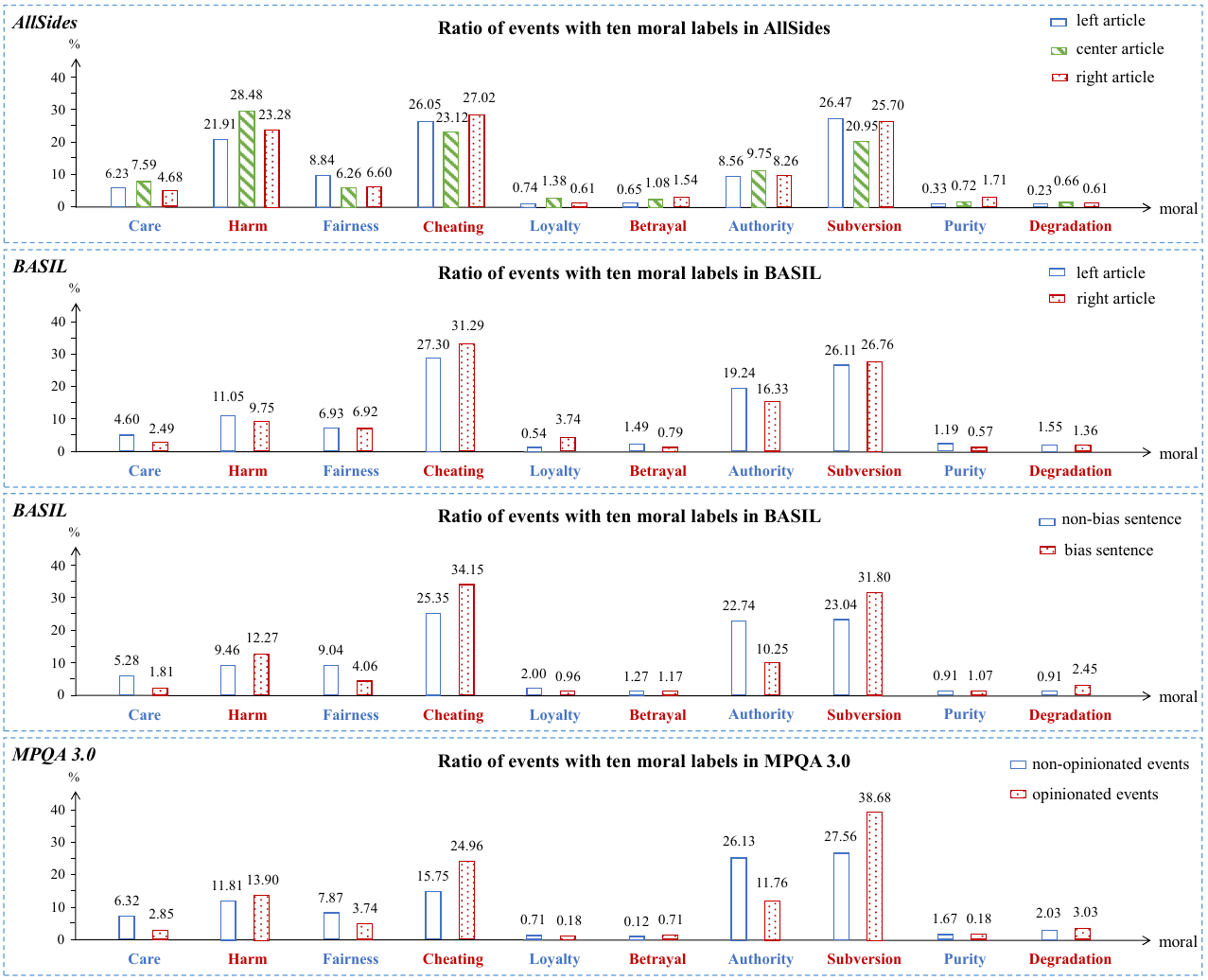}
  \caption{The ratio of moral events distributed within ten moral foundations in different portions of EMONA dataset.}
  \label{analysis_figure_2}
\end{figure*}

Figure \ref{analysis_figure_2} presents the ratio of moral events distributed within ten moral foundations in different portions of EMONA dataset. The articles with left, center, or right ideology in AllSides exhibit a different preference towards model values. The bias and non-bias sentences in BASIL showcase a different distribution of ten moral classes. The opinionated and non-opinionated events in MPQA 3.0 also present a different moral values.

\end{document}